\documentclass{article}
\usepackage{spconf,amsmath,graphicx}

\usepackage[utf8]{inputenc}
\usepackage[english]{babel}
\usepackage{color}
\usepackage{url}
\usepackage{caption}
\usepackage{multirow}
\usepackage{multicol}
\usepackage{algorithm2e}
\usepackage{todonotes}

\newcommand{\en}{\enspace} 
 
\title{Efficient video indexing for monitoring disease activity and progression in the upper gastrointestinal tract}
\name{Sharib Ali$^{\star}$ \en Jens Rittscher$^{\star}$}
\address{${\star}$Institute of Biomedical Engineering, Department of Engineering Science, \\
University of Oxford, Oxford}
\begin{document}
%
\maketitle
\begin{abstract}
\noindent{Endoscopy} is a routine imaging technique used for both diagnosis and minimally invasive surgical treatment. While the endoscopy video contains a wealth of information, tools to capture this information for the purpose of clinical reporting are rather poor. In date, endoscopists do not have any access to tools that enable them to browse the video data in an efficient and user friendly manner. 
Fast and reliable video retrieval methods could for example, allow them to review data from previous exams and therefore improve their ability to monitor disease progression. Deep learning provides new avenues of compressing and indexing video in an extremely efficient manner. In this study, we propose to use an autoencoder for efficient video compression and fast retrieval of video images. To boost the accuracy of video image retrieval and to address data variability like multi-modality and view-point changes, we propose the integration of a Siamese network. We demonstrate that our approach is competitive in retrieving images from 3 large scale videos of 3 different patients obtained against the query samples of their previous diagnosis. Quantitative validation shows that the combined approach yield an overall improvement of 5\% and 8\% over classical and variational autoencoders, respectively.
\end{abstract}
\begin{keywords}
Endoscopy, deep learning, autoencoders, Siamese network, image retrieval
\end{keywords}
\section{Introduction}
Due to the lack of efficient tools for indexing and retrieval, the wealth of information contained in endoscopy video is only rarely used for improving clinical reporting and diagnosis. Today only manually selected still frames (\textit{low quality screenshots}) are included into clinical reports. In addition, it is not feasible for endoscopists to look into  long videos. Computer assisted methods for extracting clinically relavant video frames from a larger video stream do not exist. Compressing and indexing the entire video in an extremely efficient manner would open up the possibility to use corresponding data from previous exams to enable the monitoring of disease progression and response to therapy. Content-based image retrieval (CBIR) methods allow to locate images of interest (query) in target databases. Such techniques have been already applied on large scale medical databases~\cite{Murala2012:JMS,Rodriguez:13MIR,Ye:IJCARS2017}.\\
\indent{Oesophageal} carcinoma is the sixth leading cause of mortality and the eight most common cancer worldwide. The overall 5-year survival of patients with
oesophageal carcinoma ranges from 15\% to 25\%; while earlier diagnosis can result in improved survival rate~\cite{Pennathur13:Lancet}. {Endoscopy} is a routine clinical procedure for both diagnosis and early treatment of pre-cancerous malignancies seen in oesophagus commonly referred as ``\textit{Barrett's  Oesophagus}'' (BE). Patients with BE have a 30–125-fold increased lifetime risk of developing oesophageal cancer. Barrett’s is defined as the substitution of the normal stratified squamous epithelium of the oesophagus with a columnar cell lining and can be visualized using an endoscope. It is therefore required that patients with BE must undergo periodic endoscopies during which biopsies are taken to examine the cancer risk. The goal of this work is to develop an approach that can effectively support monitoring of these patients utilizing information content of video endoscopy at an \textit{optimal compression, speed and retrieval accuracy} which are important factors for clinical usability.
\section{Related work}
Existing CBIR approaches are typically not suitable for clinical use as they utilise a very low resolution representation for the video data. Important diagnostically relevant detail is often lost. Such system use derived features such as texture, colour, shape as well as local spatial properties to learn a low-dimensional representation.~Often binary coding or hashing are used. Query images are then compared with low-dimensional representation of target images for fast image retrieval.~The efficiency of CBIR directly depends on the used feature extraction and representation. Liu {\it et al.}~\cite{LIU:PR13} used colour difference histogram utilizing colours and edge orientations for better feature representation. Murala {\it et al.}~\cite{Murala2012:JMS} employed local binary patterns (LBP) to extract features from low-level wavelet sub-bands in CT and MRI data. Scale invariant feature transform (SIFT) were used in~\cite{Rodriguez:13MIR} to derive representation for bag-of-visual words. Ye {\it et al.}~\cite{Ye:IJCARS2017} used LBP based 496-{\it dimensional} image histogram descriptor and a hashing technique based on random forest for real time biopsy retargeting utilizing video endoscopy. All of these previous approaches are based on hand-crafted features and do not incorporate high-level semantic ({\it semantic gap}).\\  
\indent{Recently,} approaches that make use of advances in deep learning have demonstrated considerable success in learning  both low- and high-level semantically relevant features. Ahmad {\it et al.}~\cite{Ahmad:JMS17} suggested using a convolutional neural network (CNN) based salient features to retrieve endoscopic images. Convolutional kernels from the first layer of a pre-trained AlexNet model was used along with a pooling strategy for achieving compact 96 bin histogram. We argue that variations in endoscopy data can not be captured using such pre-trained models as the features in endocopic images are very different from natural images (used by such pre-trained networks). Masci  and collegues~\cite{Masci:ICANN11} demontrated that unsupervised convolution autoencoders can be used to extract salient features. Krizhevsky and Hinton~\cite{Krizhevsky:ESANN11} revealed that deep autoencoders can be applied for extremely fast image retrieval task that is independent of database sizes due to high data compression capability of autoencoders. Stacked autoencoders were used by Sharma and colleagues~\cite{Sharma:AVC16} on medical images (x-ray data).\\
\indent{Two} challenges need to be addressed before image automated retrieval from endoscopy video can be applied in the clinical setting: 1) Efficient compression - the number of images per video is  extremely large (nearly 15-40 thousands) which demands for large storage (nearly 1.5-4 GB per video) and 2) preservation of diagnostically relevant features - while it is necessary to preserve a high level of anatomical detail, various artefacts need to be discarded. We argue that autoencoders are extremely well suited to address the first problem.  However, in the presence of challenges posed by data variabilities, autoencoders can fail to accurately retrieve images in restricted search space. In this paper, we propose to use autoencoders (AE) and Siamese network working side-by-side for achieving better compression, and fast and accurate image retrieval. {We} demonstrated that combining classical AE with Siamese network or variational autoencoder (VAE,~\cite{Diederik:13CoRR}) with Siamese results in improved accuracy.~We observed that VAE compresses data nearly 70 folds more than classical AE at an improved retrieval speed but with a compromise in accuracy. However, when combined with Siamese network, it yields very competitive retrieval results. We have compared retrieval performances with and without Siamese network for both autoencoders on 3 different BE patient videos. {The} rest of the paper is organized as follows: Section~\ref{sec:Method} briefly describes both classical and variational autoencoders, Siamese networks and our combined approach for image retrieval task. Section~\ref{sec:exp} evaluates our combined approaches utilizing oesophageal endoscopic video data and Section~\ref{sec:concl} concludes the paper.
\vspace{-0.35cm}
\section{Method}{\label{sec:Method}}
\begin{table*}[t!]
\centering
\begin{tabular}{lllllllllllllllllllllllll}
\hline
\multirow{2}{*}{{\bf Method}} & \multicolumn{9}{c}{{\bf Training}}   & \multicolumn{9}{c}{{\bf Compression}}  & \multicolumn{6}{c}{{\bf Testing(s)}}        \\                                                
\cline{2-9}
                        & \multicolumn{3}{c}{{\bf Samples}} & \multicolumn{3}{c}{{\bf Epochs}} & \multicolumn{3}{c}{{\bf Time(s)}} & \multicolumn{3}{c}{{\bf Data}} & \multicolumn{3}{c}{{\bf Comp.}} & \multicolumn{3}{c}{{\bf Time(s)}} & \multicolumn{3}{c}{{\bf Load}} & \multicolumn{3}{c}{{\bf Execute}} \\
                        \hline
AE                      & \multicolumn{3}{l}{33k/15k} & \multicolumn{3}{l}{500}    & \multicolumn{3}{l}{18/epoch}   & \multicolumn{3}{l}{15k} & \multicolumn{3}{l}{160~MB}    & \multicolumn{3}{l}{ 1.63}& \multicolumn{3}{l}{3}    & \multicolumn{3}{l}{0.16}\\
VAE                     & \multicolumn{3}{l}{33k/15k} & \multicolumn{3}{l}{500}    & \multicolumn{3}{l}{21/epoch}  & \multicolumn{3}{l}{15k} & \multicolumn{3}{l}{1.7~MB}    & \multicolumn{3}{l}{1.21} &\multicolumn{3}{l}{2}    & \multicolumn{3}{l}{0.18} \\
Siamese                 & \multicolumn{3}{l}{900/100} & \multicolumn{3}{l}{1000}   & \multicolumn{3}{l}{48/epoch}     & \multicolumn{3}{l}{-} & \multicolumn{3}{l}{-}   & \multicolumn{3}{l}{-} & \multicolumn{3}{l}{1.05}    & \multicolumn{3}{l}{1.35}\\     
\hline
\end{tabular}
\caption{{\bf{Details on training, test and compression.} \it All the computations were done on NVIDIA Tesla P100. Information regarding compression is provided for 15000 video images each of size $512\times 512$ and does not includes data loading time. Test times are provided for listing 100 retrieved images from a pool of 15,000 images for autoencoders and retrieval of 10 images from 100 images (each rescaled to 1$00\times 100$ pixels) for Siamese network.~\label{tab:details} }}
\end{table*}
\begin{figure}[t!]
\includegraphics[trim=0.8cm 0.25cm 0.5cm 0.0cm, clip=true,scale=0.28]{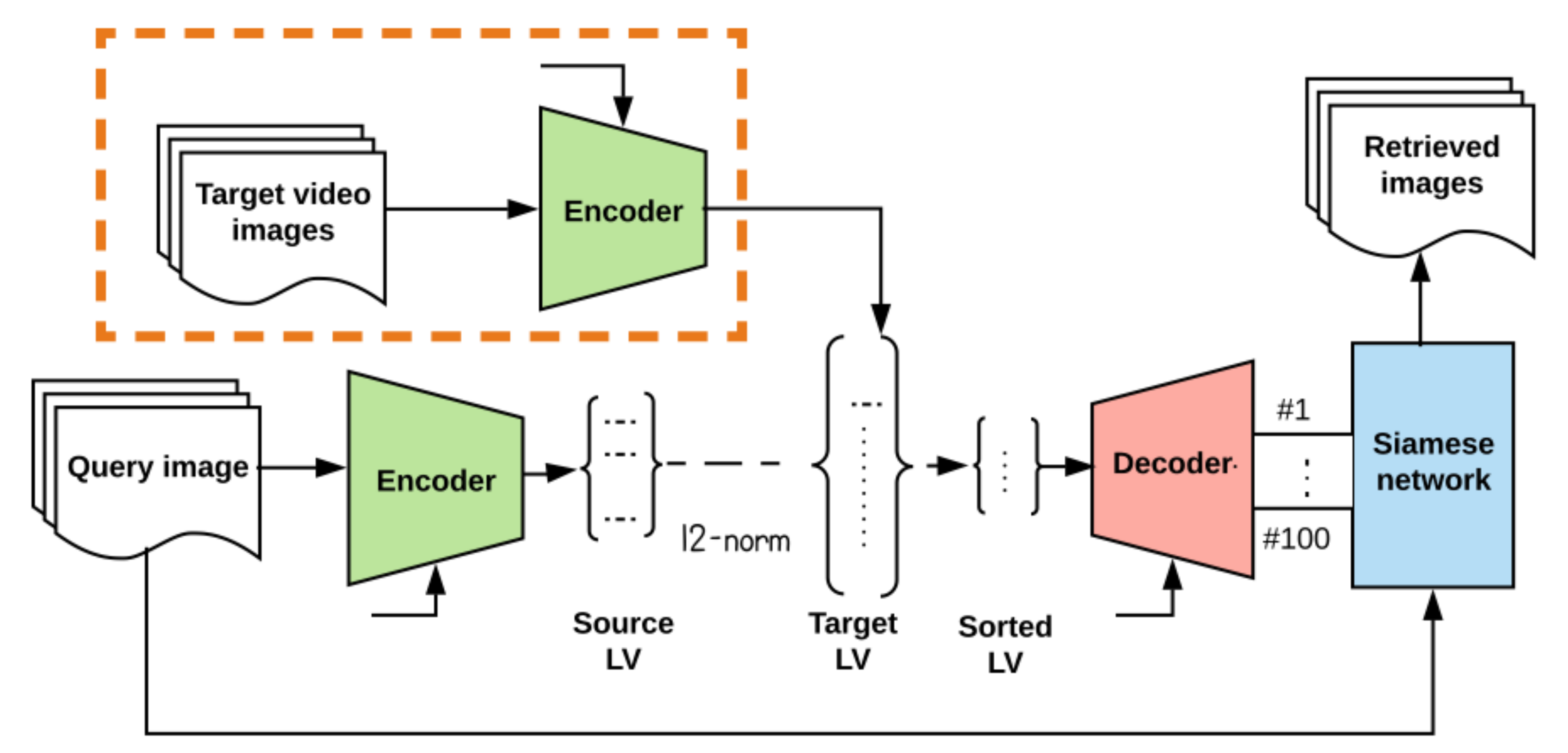}
\caption{{\bf{Autoencoder for video  frame retrieval.} \it Query and target video images are first transformed to latent variable (LV) space using trained encoders. $100$ sorted LVs based on estimated distance ($l2$-norm) between query and target LVs  are used to retrieve candidate frames from the target video. A Siamese network is used to further penalize the dissimilarity between the query and the retrieved images. Target video images are encoded (compressed) offline (dashed rectangle) to achieve real time image retrieval.}
\label{fig:blockCombinedApproach}}
\end{figure}
After providing a brief description of autoencoders and Siamese networks we motivate on why these should be combined and present our approach for efficient retrieval of video endoscopy.
%
\subsection{Autoencoder}{\label{subsec:AE}}
An autoencoder is an unsupervised machine learning algorithm that is capable of learning efficient and compressed data representation referred as ``coding`` or ``latent-space representation``, say $z$.  A reverse process ``decoding`` is performed to achieve outputs similar to the input data. Decoder tries to reconstruct using fewer number of bits from the bottleneck (latent-space). A latent-space representation is learned when the dissimilarity between decoder output and input data is minimized. \\
\indent{Here}, a convolutional autoencoder (CAE,~\cite{Masci:ICANN11}) that can be trained in an end-to-end fashion is used. Our architecture consists of 3 convolution filter layers with `relu` activation function and a fully connected last layer (32 dense connections). A subsequent downsampling with stride 2 is performed at each layer for {\it encoder} and a similar architecture with upsampling with stride 2 is performed for {\it decoder}. The decoder unflattens the encoder output first and then upscales it using similar size ({\it mirrored}) convolution filters. Cross-entropy has been used as a loss function with an Adadelta optimizer and a relu activation. The image size used for training is $124\times 124\times 3$. There are in total of 4,089,283 trainable parameters. We have trained our CAE for 500 epochs utilizing 33,000 samples for training and 15,000 for validation. 
\subsection{Variational autoencoder}{\label{subsec:VAE}}
In contrast to classical autoencoders where the information regarding input data distribution is not known, variational autoencoders~\cite{Diederik:13CoRR} assert the latent-space representation to be drawn from a unit normal distribution, $\mathcal{N}(0, {\bf I})$. Thus, such autoencoders are effective generative models that can produce samples from the learned unit normal distribution. For our image retrieval task, the encoded target embedding might not exactly match the encoded query but VAEs have tremendous strength to learn more meaningful latent representations yielding in an effective search space.\\
\indent{We} have used the same architecture (refer Sec.\ref{subsec:AE}) for our encoder-decoder network in VAE. However, the final layer of VAE encoding consists of mean $\mu$ and standard deviation $\sigma$ encoding (vectors) for 'n' ($=10$ for our case) latent embeddings. The actual coding $z$ is then randomly sampled from a unit normal distribution for decoding. Negative of cross-entropy loss is minimized using an Adam optimizer. In order to push the autoencoder to learn unit normal distribution, a second loss ``latent loss`` is used which is computed as KL-divergence between the target normal distribution and the actual coding. The total trainable parameters are 1,362,480 which is lot less than classical autoencoder as only mean and variances in the data are learnt. In our retrieval task, we compare only mean embeddings between the encoded target dataset and query images. This reduces both the computational complexity and embedding file size (see Tab.\ref{tab:details}). We have used a 10-dimensional latent variable space and trained our VAE for 500 epochs utilizing 33,000 samples for training and 15,000 for validation. 
\subsection{Siamese network}{\label{subsec:Siamese}}
A Siamese network~\cite{Koch:ICML15} learns to differentiate between two input images and consists of two identical neural networks, say ${N}_1$ and  ${N}_2$, for doing this. The dissimilarities are computed as a contrastive loss function:
\vspace{-0.25cm}
\begin{equation}{\label{eq:contractiveLoss}}
(1-Y)\frac{1}{2} \left(D_{N}\right)^2 + Y\frac{1}{2}\lbrace{\max\left(0, m -  D_{N}\right)\rbrace}^2,
\end{equation}
where $D_{N}$ is the Euclidean distance between outputs of sister networks, $Y$ is the class label and $m>0$ is the margin value. Both sister networks ${N}_1$ and ${N}_2$ have exact same weights.\\
\indent{We} have trained a Siamese network for dealing with multi-modality and varying view-points in our oesophageal endoscopy dataset. For this we created a database consisting of 100 sets of images with 10 images each (in total 1000) that included WL, NBI and 8 different viewing angles. Paired multi-modality images were generated by using a trained domain adaptive network (refere `cycleGAN'~\cite{CycleGAN2017}). CycleGAN was trained using 300 pairs of WL and NBI images. For addressing view-point changes in endoscopy, simulated images using different rotation angles ($0^\circ\leq \theta \leq 360^\circ$) were generated. {An} Adam optimizer was used to minimize contrastive loss (see Eq.~\ref{eq:contractiveLoss}). The network was trained for 1000 epochs and 100 iterations with learning parameter of 0.005. 
\subsection{Combined approach}{\label{subsec:comb}}
%
%
An overview of the proposed combined approach is presented in Fig.~\ref{fig:blockCombinedApproach}. First, both the autoencoder and the Siamese network are trained separately (see Section~\ref{subsec:AE}-\ref{subsec:Siamese}). Then, the trained autoencoder is used to compress the target video into a low dimensional latent variable space vector (target LV). Offline batch processing can be used to encode all target videos needed for patient follow-up. Thanks to the very high compression ratios this method achieves, only a modest amount of storage is required to hold the compressed videos. Given a new query image, the trained autoencoder is used to project it into the latent space. The resulting latent variable space vector (source LV) is then compared with the target LV using an $L_2$ metric. 100 LVs are sorted based on their similarity scores and are processed to either a decoder or an image list. The trained Siamese network is then used to penalise any images which don't satisfy the similarity requirement. The distance output of the Siamese network is used to produce the final ranking of the candidate images. Finally, the $n$-best retrieved images (in our case $n=10$) are being presented.
%
\section{Experiments}{\label{sec:exp}}
10 oesophageal endoscopy videos were used in this study. Each video consists of more than 15,000 frames. 4 different patient video images were combined randomly and used in different proportions for training purposes of our networks (see Table~\ref{tab:details}) and 6 videos of 3 different patients were used for our frame retrieval test (3 for query and 3 for retrieval).
\begin{table*}[t!h!]
\centering
\begin{tabular}{l|llllllllllll}
\hline
\multirow{3}{*}{{\bf Method}} & \multicolumn{9}{c}{{\bf Patient}}             & \multicolumn{3}{c}{{\bf Average}}                                     \\
                        & \multicolumn{3}{c|}{\bf \#1} & \multicolumn{3}{c|}{\bf \#2} & \multicolumn{3}{c|}{\bf \#3}& \multicolumn{3}{c}{} \\ 
               \cline{2-13}        
                        & {\bf TP}     & {\bf FP}    & \multicolumn{1}{c|}{\bf P}  & \multicolumn{1}{c}{\bf TP}    & {\bf FP}    & \multicolumn{1}{c|}{\bf P}    & {\bf TP}    & {\bf FP}    & \multicolumn{1}{c|}{\bf P}    & {\bf TP}    & {\bf FP }   & \multicolumn{1}{c}{\bf P}  \\
                        \hline
VAE                     & 356    & 134   & \multicolumn{1}{c|}{0.73}   &  351   & 139      &  \multicolumn{1}{c|}{0.71}      &  437     &   53    &  \multicolumn{1}{c|}{0.89}   &  381.3     &   108    &  0.77   \\
\hline
AE                      & 417    & 73    & \multicolumn{1}{c|}{0.85}   &  393       &    97   &    \multicolumn{1}{c|}{0.80}   &  456      &   34    &   \multicolumn{1}{c|}{0.93}  &  422     &   68    &  0.86     \\
\hline
VAE-Siamese                     & 399    & 91   & \multicolumn{1}{c|}{0.81}   &  410     &  80     &  \multicolumn{1}{c|}{0.83}      &  439     &   51    &   \multicolumn{1}{c|}{0.89 } &  416     &   74    &  0.85   \\
\hline
AE -Siamese                    & 437   & 53   & \multicolumn{1}{c|}{0.89}   &   437    &  53     &  \multicolumn{1}{c|}{0.89}      & 473      &   17    &   \multicolumn{1}{c|}{0.96}   &  449     &   41    &  0.91  \\
\hline
\end{tabular}
\caption{{\bf{Query frame retrieval for BE diagnosis/treatment:} \it Average TP (true positive), FP (false positive) and P (precision) are provided for 49 query samples randomly selected from previous visits of each patient video endoscopy. All values were calculated for 10 first retrieved images for each case with reference to the query samples by an expert.~\label{tab:quantitative}}}
\end{table*}
\vspace{-0.35cm}
\subsection{Quantitative results}
\begin{figure}[t!]
\centering
\begin{minipage}[b]{0.20\linewidth}
\includegraphics[scale=0.3]{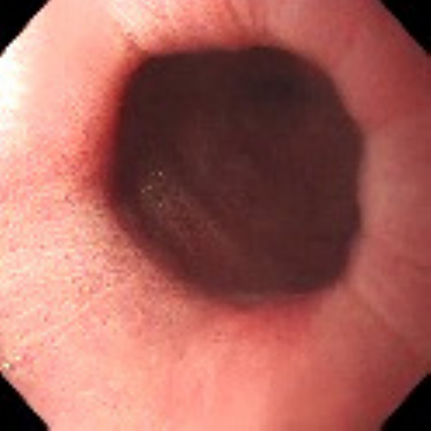}
\centerline{Query  \#1}\medskip
\end{minipage}
\begin{minipage}[b]{0.20\linewidth}
\includegraphics[scale=0.3]{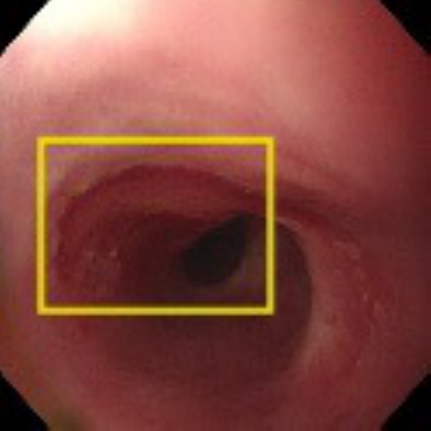}
\centerline{Query \#2}\medskip
\end{minipage}

\begin{minipage}[b]{0.99\linewidth}
\includegraphics[scale=0.1]{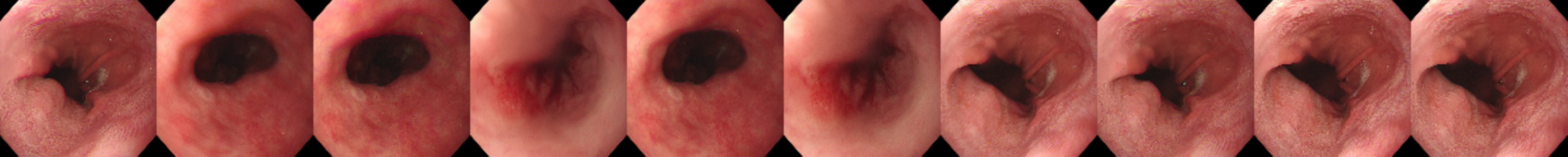}
\end{minipage}

\begin{minipage}[b]{0.99\linewidth}
\includegraphics[scale=0.1]{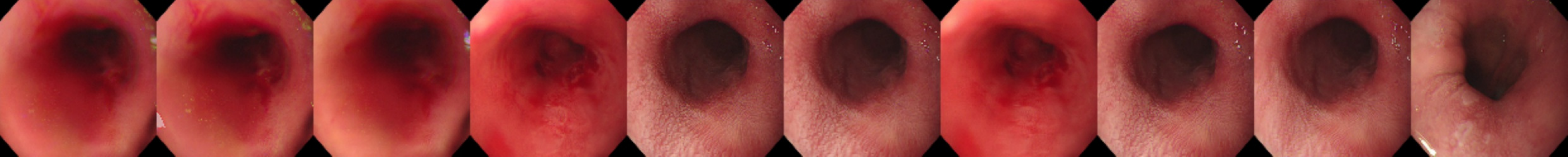}
\end{minipage}

\begin{minipage}[b]{0.99\linewidth}
\includegraphics[scale=0.1]{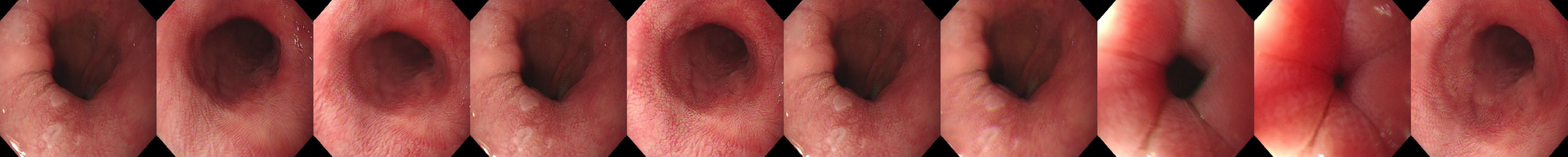}
\end{minipage}

\begin{minipage}[b]{0.99\linewidth}
\includegraphics[scale=0.1]{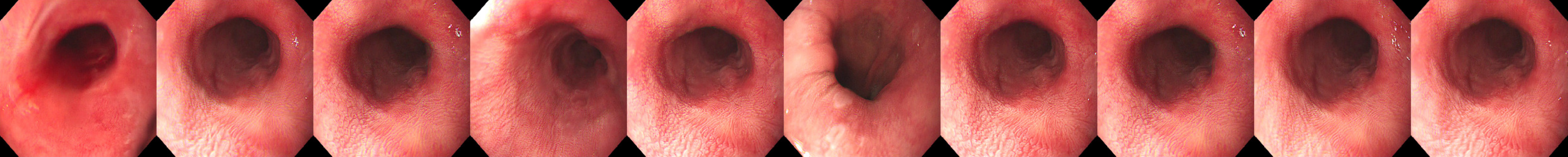}
\end{minipage}

\vspace{2mm}

%
%
%
%

\begin{minipage}[b]{0.99\linewidth}
\includegraphics[scale=0.1]{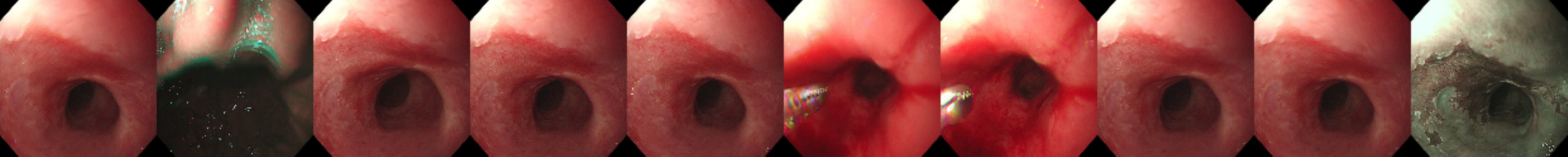}
\end{minipage}

\begin{minipage}[b]{0.99\linewidth}
\includegraphics[scale=0.1]{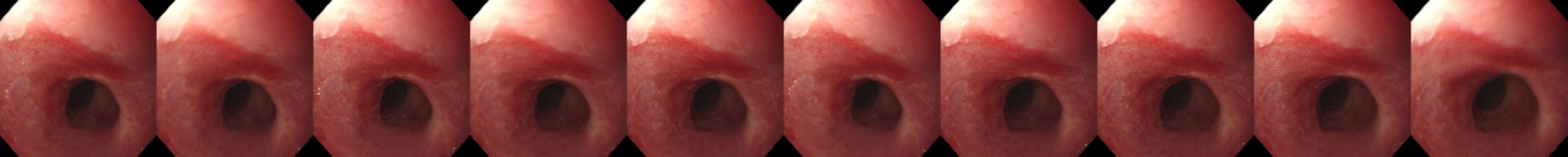}
\end{minipage}

\begin{minipage}[b]{0.99\linewidth}
\includegraphics[scale=0.1]{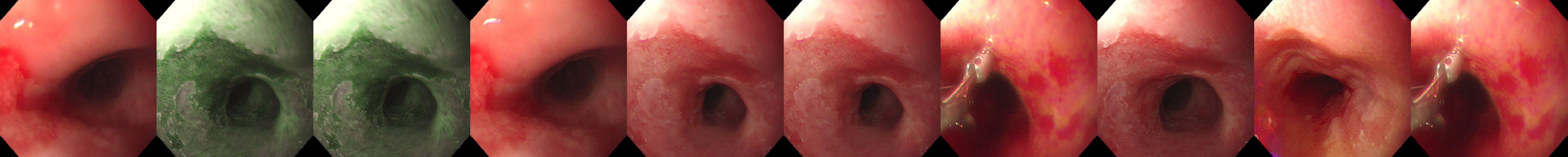}
\end{minipage}

\begin{minipage}[b]{0.99\linewidth}
\includegraphics[scale=0.1]{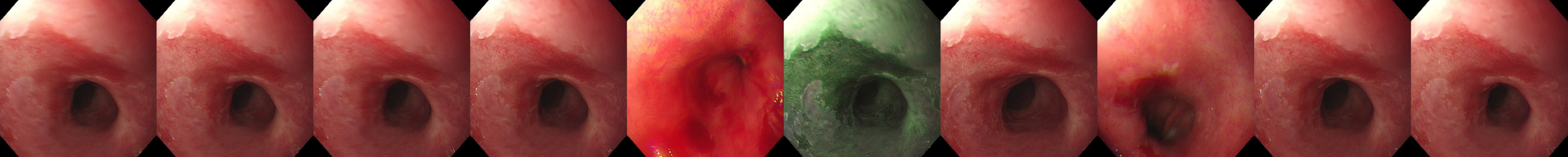}
\end{minipage}

\caption{{\bf{Visual analysis of improved performance of proposed approach.~}\it Query \#1 and query \#2 resepectively represents upper esophageal sphincher and a Barrett's region (marked by yellow rectangle) both of which are crucial landmarks for endoscopists. Retrieval results for \#1 and \#2 are shown in top and down blocks, respectively. In each block, 1st row: VAE, 2nd row: AE, 3rd row: VAE-Siamese and 4th row: AE-Siamese. It can be observed that our combined approach (i.e., last two rows of each block) is able to retrieve the most similar images incorporating more variabilities than when autoencoders are used alone.}\label{fig:qualititativeResults}}
\end{figure}
Table~\ref{tab:quantitative} shows image retrieval performance of both the sole application of autoencoders and our combined approaches. It can be seen that Siamese network improves the average retrieval  precision by 8\% and 5\%, respectively, for VAE and AE. AE-Siamese yields the best results  for all 3 patient cases (89\%, 89\% and 96\%) and on average 6\% higher precision than VAE-Siamese. However, from Tab.~\ref{tab:details} it can be observed that VAE has the best compression performance, nearly 70 folds more than AE, and a very reasonable average precision of 85\% (see Tab.~\ref{tab:quantitative}).
%
 %
 \vspace{-0.2cm}
\subsection{Qualitative results}
Fig.~\ref{fig:qualititativeResults} presents visual analysis of image retrieval on two query video images in our dataset. These query images were searched in a same patient video archived 6 months earlier and consisted of nearly 21,462 image frames. For query \#1 (top block), VAE (1st row) has more mismatches than AE (2nd row) and the retrieved images in both cases are not ordered even if they are matched. We can observe that VAE-Siamese (3rd row) has improved matches while AE-Siamese (4th row) has also been ordered better. Similarly, for query \#2 (bottom block), AE (2nd row) has better matches than VAE (1st row). Infact, AE has perfect match for same modality. However, utilizing Siamese network on top, both VAE and AE (3rd and 4th rows, respectively) are able to capture more variabilities for the same site and includes NBI multi-modality cases.
\vspace{-0.1cm}
\section{Conclusion}{\label{sec:concl}}
While autoencoders allow for better compression of large-scale endoscopy data eliminating the need for large archival spaces, our experiments demonstrated that a Siamese network on top can be used to provide a very effective similarity score for improved retrieval accuracy of clinically significant frames. Our resulting system can thus achieve high compression and maintain a feature representation that keeps diagnostically relevant detail in the context of monitoring Barrett's oesophagus. Our future work will include combining of text information on the report with image data for obtaining more accurate and meaningful image retrieval of oesophageal endoscopic videos.
%
\vspace{-0.25cm}
\section*{Acknowledgement}
\vspace{-0.25cm}
SA is supported by the NIHR Oxford BRC. JR is funded by EPSRC EP/M013774/1 Seebibyte. 
\bibliographystyle{IEEEbib}
\bibliography{bib_endoscopy}

\end{document}